# TLRNet: Estimating Individual Treatment Effect based on Local Information and Single Learner Structure

Ali Haghpanah Jahromi
*Department of Electrical and Computer Engineering, University of Shiraz*
*Shiraz, Iran*
*alihgpj@gmail.com*

Mohammad Taheri
*Department of Electrical and Computer Engineering, University of Shiraz*
*Shiraz, Iran*
*motaheri@shirazu.ac.ir*

Zohreh Azimifar
*Department of Electrical and Computer Engineering, University of Shiraz*
*Shiraz, Iran*
*azimifar@cse.shirazu.ac.ir*

***Abstract*— Causal inference has become a central issue across various fields, including computer science, statistics, economics, education, healthcare, and medicine. The broad applicability of this discipline has garnered increased research funding and attention. In recent years, the estimation of causal effects from observational data has gained traction due to the vast amounts of collected data and the lower costs compared to randomized controlled trials. Advances in causal effect estimation methods have enhanced service personalization tools. For instance, these tools can help identify the most effective type of treatment (considering both cost and success rate) for each patient among different medical service options. This paper proposes an innovative method for estimating the heterogeneity of treatment effects. The structure of the proposed model is based on a deep neural network and a pseudo-single learner. The proposed method has been compared with other state-of-the-art methods on the IHDP benchmark. Acceptable results have been obtained by using one estimator to estimate the potential outcomes of two treatment groups. Accordingly, this paves the way for further development and improvement of the proposed method.**

## I. INTRODUCTION

Causality inference modeling refers to the modeling of cause-and-effect relationships in phenomena. Although causality and correlation are sometimes used interchangeably in everyday language, these two words have different interpretations in the scientific literature of causality inference. Correlation describes the increasing or decreasing trend of two variables simultaneously. In the discussion of causality inference, the cause is partly responsible for the occurrence of the effect, and the effect partly relies on the cause. The causality inference aims to investigate the changes in the effect variable when applying modifications to the cause variable [1, 2, 3].

Various structures and frameworks have been introduced to model causality inference, including a Structural Causal Model (SCM) [4, 5] and the Neyman–Rubin Potential Outcomes or the Rubin Causal Model (RCM). This paper's proposed method is based on the Neyman-Rubin potential outcomes framework, which evaluates the treatment variable's causality effect on the problem's outcome under confounding factors [6, 7]. The presence of confounding factors and selection bias are the central challenges to causality analysis and estimating treatment effect heterogeneity from observation data. Thus, several methods have been presented to overwhelm these problems, each of which has studied various aspects and capabilities of problem-modeling structures [1].

### A. Applications

Heterogeneous treatment effect estimation is commonly used in various contexts. For instance, one study noted that drug effects can vary among individuals and that various confounding factors can influence drug effectiveness. Consequently, researchers have examined methods for estimating heterogeneous causal effects [8]. Considering the importance of studying interventions in actionable healthcare, assessing the causal effect is essential for analyzing different scenarios of medical interventions [9]. It should be noted that Randomized Clinical Trials (RCTs) are not appropriate for studying the estimation of individual treatment effects. For example, while one type of treatment may be suitable for one COVID-19 patient with lung disease, the same kind of treatment may endanger the life of another COVID-19 patient with another condition. For this reason, analyzing the individual treatment effect on the patient is essential in personalizing the therapy method for patients [10].

Today, advertising campaigns are used in various business programs, and their effectiveness is one of the most crucial marketing questions. For illustration, does an advertising campaign lead to an increase in the number of clicks or sales? The high cost and time-consuming nature of random experiments have attracted the attention of commercial and industrial activists to the use of observational data for estimating the effectiveness of advertising campaigns [11, 12, 13, 14]. Providing suitable suggestions to users can help them with various issues. Recommendation systems can offer valuable recommendations to users using causal effect estimation. The high volume of observational data in recommendation systems encourages businesses to employ causal effect estimation tools. In these systems, selection bias and confounding factors are inherent in the data because users' biased choices are expected [15, 16].

### B. *Literature review*

In this article, the Neyman-Rubin potential outcome framework is used. For each unit of the feature vector (known as covariate or cofounder) $x \in X$, action $t \in \{1,0\}$ (also known as treatment or intervention), and two potential output values $Y(1)$, and $Y(0)$ corresponding to each treatment are defined. According to the principle of causality inference based on the Neyman-Rubin potential framework, only one of the potential outcomes occurs in the observation data for each unit, which is called the consistency assumption: if t=1 the y=Y(1) is observed, and if t=0 the y=Y(0) is observed [1, 17]. To better understand this topic, consider a study on the effectiveness of a new blood pressure medication on several patients. Depending on whether they receive the medication, two possible actions can be taken for each patient. Patients are divided into two treatment groups: the treatment group (t=1), which receives the medication, and the control group (t=0), which does not. The medical records of each patient provide pre-treatment characteristics that remain unchanged regardless of the treatment type but may influence the assignment of treatments to patients. For example, a research expert may determine that older adults are more suitable candidates for receiving blood pressure medication. According to the example, the measurement output for each patient is their blood pressure. Only one of the potential outcomes (Y(1) or Y(0)) is visible among the potential outcomes for the two possible actions. Because only one of the potential outcomes is observed in the observational data, the causal effect problem is estimating the counterfactual potential outcomes for each patient [3].

The difference (Y(1)-Y(0)) or ratio (Y(1)/Y(0)) of potential outcomes is usually used to calculate the treatment effect estimand. In the mentioned example, to estimate the treatment effect with potential outcomes, the difference between the outcomes of the treatment group and the control group can be employed as the treatment effect. If this value is greater than zero, it means that the use of the drug did not lead to a decrease in blood pressure, and If it is less than zero, it means that the drug is effective in reducing blood pressure. In the inference of causality and treatment effect estimation, two parameters of Average Treatment Effect (ATE) and Conditional Average Treatment Effect (CATE) are usually considered for analysis, as described below [3, 17].

$$ATE = \mathbb{E}[Y(1) - Y(0)] \tag{1}$$

$$\widehat{ATE} = \frac{1}{N}\sum_{i=1}^{N}(y_i(1) - y_i(0)) \tag{2}$$

$$\begin{aligned} \tau(x) &= m_1(x) - m_0(x) \\ &= \mathbb{E}[Y(1|X = x)] - \mathbb{E}[Y(0|X = x)] \\ &= \mathbb{E}[Y(1|X = x) - Y(0|X = x)] \end{aligned} \tag{3}$$

In this context, $m_1(x)$ and $m_0(x)$ represent functions based on the potential outcomes for a feature vector x. The term $\tau(x)$ denotes the expected treatment effect when the treatment is applied (t=1) compared to when it is not applied (t=0) for an individual with feature vector x. The $\tau(x)$ is called the Individual Treatment Effect (ITE) or CATE. The causal effect objective is to estimate $\hat{\tau}(x)$ as an approximation of $\tau(x)$ in such a way that a loss function, such as square loss of $\hat{\tau}(x)$ and $\tau(x)$, remains small.

Three fundamental assumptions are essential for estimating treatment effects. The first assumption is that no hidden covariates exist, meaning that experts fully identify all relevant pre-treatment variables. The second assumption is that the potential outcomes are independent of the treatment assignment mechanism, given the pre-treatment variables $((Y(1),Y(0)) \perp t|x)$. The third assumption pertains to the probability of assigning any unit to the treatment group, which must always be positive $(0<p(t=1 \mid x)<1)$. In the literature on causal inference, this probability is commonly called the propensity score.

According to the principles of causal inference, it is not possible to simultaneously observe the outcomes of different actions on the same individual. For instance, a person cannot simultaneously be part of the treatment group (receiving medication) or the control group (not receiving medication). Further, in the causal inference literature, a unit at different time points is considered a different sample. Since observational data only provide one potential outcome, it is impossible to utilize conventional machine learning methods to estimate heterogeneous treatment effects directly.

The proposed method is based on a neural network structure that estimates the potential outcomes for each treatment group, allowing for the calculation of treatment effect estimates. Below is a summary of the paper's contributions.

- The proposed method uses different representation functions to extract the local information obtainable in the treatment and control groups, which is not seen in previous methods.
- Unlike previous techniques, which use two separate estimators for the treatment and control groups, the proposed method presented in this article employs a single estimator. The use of an estimator improves the chances of the model's generalizability due to the use of all the samples of both groups, but of course, it also creates challenges.

The paper is structured into three sections. The second section outlines the activities in this field. The third section introduces the proposed method, a model designed to estimate heterogeneous treatment effects. The fourth section presents the experimental results and evaluation of the proposed method, and the conclusion succinctly summarizes the findings.

## II. RELATED WORK

Various methods have been developed to estimate the causal treatment effect using decision trees, neural networks, Bayesian networks, and statistical regression models.

### A. *Tree based method*

The BART (Bayesian Additive Regression Trees) method employs a collection of trees as weak learners to create a regression model [18]. In this context, let W represent a binary tree with interior and terminal nodes, along with a set of parameters (M) related to the terminal nodes of the tree. The BART structure evaluates the output value using functions derived from these weak trees [1, 19]. The BART model can be expressed with the following equation:

$$Y = g(x; W_1, M_1) + g(x; W_2, M_2) + \cdots + g(x; W_m, M_m) + \varepsilon \quad where\ \varepsilon \sim \mathcal{N}(0, \sigma^2) \tag{4}$$

One of the advantages of the BART method is its comfort of implementation. Because the estimation is non-parametric, there is no need to rely on prior information about the input and output variables. BART is used to estimate both the CATE and the ATE. Propensity score matching, weighting, and regression adjustment can also enhance its estimations [19]. Causal trees are another decision tree structure developed to estimate the treatment effect. They create an Honest tree by dividing the data into two categories, one for estimating the decision tree and the other for estimating the treatment effect. The developed version of these trees is also presented under the Causal Forest, which uses more than one tree for estimation [20, 21, 22].

### B. Meta Learners

A group of methods known as meta-learners are developing to estimate treatment effects based on two key factors. First, it is essential to eliminate the spurious correlation between confounders and the target variable (The $m_1(x)=E[Y(1)|x]$ and $m_0(x)=E[Y(0)|x]$ are estimated by the ($\widehat{m}_1(x)$ and $\widehat{m}_0(x)$). Second, an accurate estimator must be trained to estimate the CATE. Meta-learning methods consider both factors, while many other methods do not. Generally, meta-learning models follow a two-step procedure. The first step involves training basic learning models to estimate the outcomes for each treatment group. In the second step, a conditional treatment effect estimator is created based on the results from the first step. Several meta-learning methods have been introduced, including T-learner [23], S-learner [23], X-learner [23], U-learner [24], and R-learner [24].

In the S-learner approach, the treatment feature is concatenated with the pre-treatment features, and a regression model is trained on these combined features to estimate the target variable ( $\widehat{m}(X=x, T=t) = \hat{f}([X=x, T=t])$ ). Then, the treatment feature is set to "1" during the treatment group mode to estimate the treatment potential outcome. Similarly, the treatment feature is set to "0" for the control group to estimate control potential outcomes [1, 23].

In the T-Learner method, two separate models ($\widehat{m}_1(X=x) \sim E[Y(1)|x]$ and $m_0(X=x) \sim E[Y(0)|x]$) are used to estimate the outcomes for each treatment group. The treatments' variable help differentiate the samples for feeding these two models. Each model is trained to predict the potential outcomes for its respective treatment group, which is necessary for estimating the treatment effect. Estimating the treatment effect using the T-Learner and S-Learner methods is illustrated below [1, 23].

$$\begin{aligned} \hat{\tau}_{S-Learner}(x) &= \widehat{m}(X = x, T = 1) - \widehat{m}(X = x, T = 0) \\ \hat{\tau}_{T-Learner}(x) &= \widehat{m}_1(X = x) - \widehat{m}_0(X = x) \end{aligned} \tag{5}$$

The X-Learner is a meta-learning method that uses the T-Learner model to estimate the response functions for each group separately (equation 5). It calculates the treatment effect for each sample based on the relationship defined in the following equation 6.

$$\begin{aligned} \mu_1(x) &= E(Y(1)|X = x) \\ \mu_0(x) &= E(Y(0)|X = x) \end{aligned} \tag{6}$$

$$\begin{aligned} \widetilde{D}_i^1 &:= Y_i^1 - \hat{\mu}_0(X_i^1) \\ \widetilde{D}_i^0 &:= \hat{\mu}_1(X_i^0) - Y_i^0 \end{aligned} \tag{7}$$

In the next step, two CATE models are trained on treatment effects that have previously been obtained. The $\hat{\tau}_1(x)$ is trained to estimate the causal effect for the treatment group based on the results from the previous stage ($\hat{\tau}_1(x)=E(\widetilde{D}_i^1|X=x)$). In contrast, the $\hat{\tau}_0(x)$ is trained to estimate the treatment's causal effect for the control group ( $\hat{\tau}_0(x) = E(\widetilde{D}_i^0|X=x)$ ). The final model for estimating the treatment effect is derived from the weighted sum of the outputs of these two CATE models. The propensity score is typically used as the weight for each treatment effect estimate [23].

$$\hat{\tau}(x) = g(x)\hat{\tau}_0(x) + \left(1 - g(x)\right)\hat{\tau}_1(x) \tag{8}$$

The R-Learner method utilizes cross-validation to develop multiple base models for estimating the response surface of each treatment group and the propensity score. By integrating these base models, a final model is constructed to estimate the heterogeneous treatment effect using the R-Learner estimator. The R-learner objective function has minimized the error of treatment effect by employing a combination of estimators. In the equation 8, $\widehat{m}^{-i}(X_i)$ and $\hat{e}^{-i}(X_i)$ represent the estimator for the surface response of treatment groups and propensity score conducted without the i-th training sample. Initially, the method estimates the $\hat{e}(x)$ and $\widehat{m}(x)$ functions using cross-validation with machine learning models. Afterward, the treatment effects are estimated by minimizing the R-loss (equation 9) [24].

$$\hat{L}_n(\tau(x)) = \frac{1}{n}\sum_{i=1}^{n}\left[\left(Y_i - \widehat{m}^{-i}(X_i)\right) - \left(T_i - \hat{e}^{-i}(X_i)\right)\tau(x)\right]^2 \tag{9}$$

The Doubly Robust Optimal Estimation method is an approach for imputing the CATE. This method divides the sample set, consisting of {Y, X, and T}, into three subsets {$Y_i$, $X_i$, and $T_i$ where i=1,2,3}. The first subset is used to estimate the propensity score ($\hat{e}^0(x)$), while the second subset is utilized to approximate $\widehat{m}_1^1$ (x) and $\widehat{m}_1^1(x)$. The CATE is then estimated using the established doubly robust relationship.

$$\hat{\tau}(x) = \frac{T - \hat{e}(x)}{\hat{e}(x)(1 - \hat{e}(x))}\left(Y - \widehat{m}_T(x)\right) + \widehat{m}_1(x) - \widehat{m}_0(x) \tag{10}$$

By thoroughly iterating through the three subsets of samples to estimate the different models for the propensity score and the response surfaces of the treatment groups, the base models are obtained from three distinct sets of models ($e^i(x)$, $m_0^i(x)$, $m_1^i(x)$) for each subset of samples. Each of these models contributes to the CATE estimate, and the final CATE is computed by averaging all the estimations, providing a comprehensive policy.

### C. Representation learning

Representation learning involves transforming input data by mapping the input space to a new feature space. This process combines the original features to create a feature space with more information. When estimating treatment effects, representation learning typically pursues two main objectives. The first objective is to balance the input domains of the treatment and control groups. Common support between the treatment and control groups can prevent estimating treatment effects from becoming an extrapolation problem. Hence, many representation methods focus on defining shift covariates that map the treatment and control samples into space with better common support or more statistical interference. The second objective of representation learning is to ensure that the extracted features effectively describe the outputs [1].

Treatment-Agnostic Representation Network (TARNet) is a representational learning method utilized to estimate CATE. It aims to create a transformation space where the outputs of the treatment and control groups are accurately estimated while ensuring common support between treatment groups. To achieve this, TARNet employs a network with two separate heads for regression estimation and contains the representation network before regression heads to provide suitable common support for sample distribution. TARNet first maps samples into a representation space. Depending on whether a sample belongs to the treatment or control group, it is processed by one of the two heads of the network for the estimation of factual outcomes. This network is trained end-to-end, and its objective function consists of two main components: the first represents the samples' regression loss, and the second measures the statistical distance between the samples from the two treatment groups. If the alpha value in this objective function is higher than zero, the network is called Counterfactual Regression Net (CFRNet).

$$\begin{aligned} \min_{\substack{h,\Phi \\ \|\Phi\|=1}} \quad & \frac{1}{n}\sum_{i=1}^{n} w_i \cdot L\left(h(\Phi(x_i), t_i), y_i\right) + \lambda \cdot \mathfrak{R}(h) \\ & + \alpha \cdot \mathrm{IPM}_G\left(\{\Phi(x_i)\}_{i:t_i=0}, \{\Phi(x_i)\}_{i:t_i=1}\right), \\ \text{with} \quad & w_i = \frac{t_i}{2u} + \frac{1-t_i}{2(1-u)}, \quad \text{where} \quad u = \frac{1}{n}\sum_{i=1}^{n} t_i, \\ \text{and} \quad & \mathfrak{R} \text{ is a model complexity term.} \end{aligned} \tag{11}$$

In the above relationship, $\Phi$ and $h$ are representative and predictive networks, respectively, and L is the regression loss function. This article also presents generalization bounds for estimating the upper bound of the precision in estimating heterogeneous effects (PEHE). This error bound is defined based on the regression loss and the distance between the two distributions of the control treatment samples (equation 12).

$$\begin{aligned} \epsilon_{PEHE}(h,\varphi) &= \int_X \left(\hat{\tau}(x) - \tau(x)\right)^2 p(x)dx \\ \leq \epsilon_F^{t=0}(h,\varphi) &+ \epsilon_F^{t=1}(h,\varphi) + B_\varphi IPM_G\left(p_\varphi^{t=0}, p_\varphi^{t=1}\right) - 2\sigma_Y^2 \end{aligned} \tag{12}$$

In the equation above, $\epsilon_F^{t=0}(h,\varphi)$ and $\epsilon_F^{t=1}(h,\varphi)$ represent the square regression losses for the treatment and control groups, respectively. The difference in distribution between these groups is measured by the integral probability metric (IPM), and $\sigma_Y^2$ refers to the variance of the outcomes.

Dragon-Net is a neural network framework similar to TARNet designed to enhance the estimation of treatment effects by adjusting propensity scores and utilizing target regularization. This structure leverages the propensity score to refine treatment effect estimations and modify the representation space. Additionally, the developed model employs a targeted regularization method that promotes optimal asymptotic properties in the models [25]. Other methods, such as GANITE, use generative networks to produce counterfactual variable values based on the observed sample distribution [26].

## III. Proposed Method

### A. Problem setup

One S-learner drawback is that the situation covariates correlate highly with the outcome variable; the modeling process may ignore the treatment variable. This issue is more likely to occur when there are many covariate features. For instance, selection bias can be inherent in observational data or may arise from the preferences of research experimenters. In a selection bias case, each covariate can affect the treatment variable differently, resulting in units with specific covariate features being assigned with significant bias to the treatment group. This information is not directly incorporated into an S-learner modeling.

### B. Treatment-Locale Representation Network (TLRNet)

The proposed method aims to establish separate mappings for treatment and control features to extract and embed relevant information about both groups in informative spaces. The informative space provides valuable information about the relationship between treatment groups and the assignment mechanism, serving as a practical description for estimating factual outcomes. For this purpose, two mapping functions: $\varphi_t$ and $\varphi_c$ are considered. The $\varphi_t$ function is responsible for extracting information related to the treatment group to estimate the treatment factual outcome, while the $\varphi_c$ function focuses on the control group. The extracted information should encapsulate the background details of each group and describe the actual outcomes. This mapping involves identifying the covariates and units that create distinctions between the two groups and integrating their information to depict the differences between the treatment groups for the single learner part.

Figure 1 illustrates the proposed network's structure. It consists of two mapping functions: one for the treatment group and another for the control group. After mapping, the samples are processed through the output prediction network to predict factual outcomes. This predictive network receives samples from both groups and functions as an pseudo S-Learner.

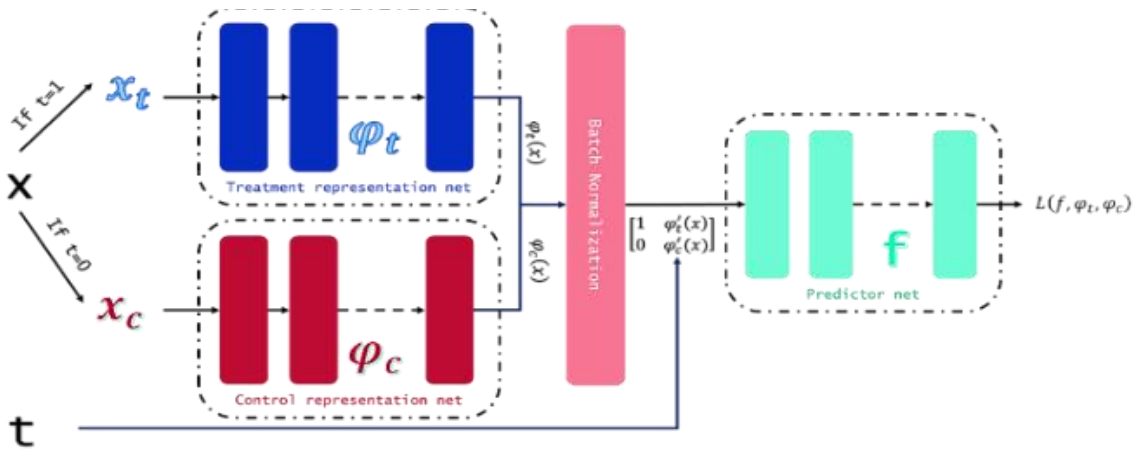

Figure 1 TLRNet Structure

It is crucial to understand that the network is trained end-to-end, a departure from the traditional approach of training different components separately. This strategy harnesses the benefits of a neural network architecture. The error gradient for samples in the treatment group is applied solely to the predictive and treatment representational networks. Similarly, the error gradient for control group samples is applied only to the predictive and control representational networks. Additionally, the treatment feature is concatenated with other features from the representation networks for better integration. We call this network the Treatment-Locale Representation Network (TLRNet). The proposed method's objective function includes regression loss function terms for factual outcomes and an adjustment term for parameter regularization.

$$\underbrace{min}_{f,\varphi_t,\varphi_c} \sum_{i=1}^{n} t_i . L\left(f\left(\varphi_{t_i}(x_i), t_i\right), y_i\right) + \gamma \left|\left|\Re(f)\right|\right| \quad (13)$$

In the above equation, $\varphi_{t_i}$ the denotes the representation function performed on unit i, according to the treatment label $t_i$ and $f(\varphi_{t_i}(x_i),t_i)$ refers the predictor network.

The loss function L(.) can be chosen from any regression loss function in which the summation of square error is employed in the experiment.

While promising, the TLRNet design can result in significant fluctuations in network convergence, a factor that requires careful consideration and management. To overcome the exploding gradient, a batch normalization layer is applied to center the samples from both groups before they are fed into the predictive network. The pseudo-code for the TLRNet algorithm is provided below.

Algorithm TLRNet:

1. **Input**: Factual sample $(x_1,t_1,y_1),\ldots,(x_n,t_n,y_n)$, Loss function L(.), Treatment representation network $\varphi_t$ with initial weights $W_t$, Control representation network $\varphi_c$ with initial weights $W_c$, Predictor network f with initial weights $W_f$
2. **While** not converged **do**
   a. Sample mini-batch $\{i_1,i_2,\ldots,i_m\}$
   b. Separate the treatment and control samples to map them with $\varphi_t$ and $\varphi_c$, respectively
   c. Perform the Batch Normalization Layer over samples
   d. Predict the factual outcomes with the predictor network and calculate the loss function L(.)
   $$g_t = \nabla_{W_t} \sum_{i=1}^{n_t} L(f(\varphi_t(x_i), 1), y_i)$$
   $$g_c = \nabla_{W_c} \sum_{i=1}^{n_c} L(f(\varphi_c(x_i), 0), y_i)$$
   e. $$g_f = \nabla_{W_f} \sum_{i=1}^{n_t} L(f(\varphi_t(x_i), 1), y_i) + \sum_{i=1}^{n_c} L(f(\varphi_c(x_i), 0), y_i)$$
   f. Update $[W_t, W_c, W_f]$ according to the $[g_t, g_c, g_f]$
   g. Check convergence criterion
3. **End while**

## IV. EXPERIMENTS

Evaluating causal inference algorithms is more complex than assessing traditional machine learning algorithms. According to the basic principles of causal inference, any individual's treatment effect is unknown, and only the potential outcome of one action is observable. Therefore, analyzing causal inference algorithms requires methods that allow access to the problem's ground truth outcomes. Simulations are conducted on real-world data to create synthetic potential outcomes for various treatment groups to overcome the absence of ground truth.

### *A. Simulated outcome: IHDP*

One well-known simulation in this field is the semi-synthetic IHDP simulation provided by Hill. This simulation was designed to estimate conditional causal effects based on the Infant Health and Development Program (IHDP) data [19, 17]. It includes covariates extracted from a randomized controlled trial that studied the impact of specialist home visits on infant cognitive education scores. Some samples were removed biasedly to introduce imbalance into the treatment group. The final dataset contains 747 samples, with 139 belonging to the treatment group and 608 to the control group. The dataset includes 25 covariate variables, and this research utilized 1,000 generated simulations with predetermined test samples. Additionally, 10% of the dataset was pre-selected as test data, while the remaining 90% was used to train models for estimating treatment effects.

### *B. Methods under comparison*

The article compares treatment effect estimation methods based on Meta-learning, Doubly Robust (DR), and neural networks like GANITE and DragonNet. The researchers evaluated some of these algorithms using available libraries for treatment effect learning. They referenced results from relevant articles listed in Table 1 for other methods. For the methods that require a base learner, gradient boosting (LightGBM) is used, which is a robust and scalable algorithm.

Table 1 Algorithms and Packages

| Method | Base Learner | Package or Paper Result Source |
|---|---|---|
| DR-Learner | LightGBM [27] | Causal ML [28] |
| S-Learner [23] | LightGBM [27] | Causal ML [28] |
| T-Learner [23] | LightGBM [27] | Causal ML [28] |
| X-Learner [23] | LightGBM [27] | Causal ML [28] |
| R-Learner [24] | LightGBM [27] | Causal ML [28] |
| DragonNet [25] | - | Causal ML [28] and Paper |
| GANITE [26] | - | Paper |
| TARNet [17] | - | Paper |
| CFRNet [17] | - | Paper |

### *C. Results*

The PEHE criterion assesses the mean squared error between the estimated treatment effect and the real CATE. This relationship is illustrated by calculating the square of this error for the training and test data, with the results shown in Table 2. The algorithms' outcomes have been reported based on 1,000 simulations of IHDP data. Table 2 shows the best results for each criterion and the results of the TLRNet method.

$$\epsilon_{PEHE} = \frac{1}{n}\sum_{i=1}^{n}\left(\hat{\tau}_i(x) - \tau_i(x)\right)^2$$
$$= \frac{1}{n}\sum_{i=1}^{n}\left(\left(f(\varphi_t(x_i),1) - f(\varphi_c(x_i),0)\right) - \left(m_1(x_i) - m_0(x_i)\right)\right)^2 \quad (14)$$

The average treatment effect is a critical parameter in causality inference studies. The following equation calculates the average error of the treatment effect. This equation computes the absolute difference between the actual and estimated average of the heterogenous treatment effect over 1,000 simulations of IHDP.

Table 2 Mean and STD for 1000 IHDP

| | IHDP √ **PEHE** | | IHDP $\epsilon_{ATE}$ | |
|---|---|---|---|---|
| Method name | Within-sample | Out-of-sample | Within-sample | Out-of-sample |
| CFRNet | 0.71±0.02 | **0.76±0.02** | 0.25±0.01 | 0.27±0.01 |
| TARNet | 0.88±0.02 | 0.95±0.02 | 0.26±0.01 | 0.28±0.01 |
| DragonNet | 0.82±0.28 | 0.92±0.50 | 0.14±0.01 | **0.20±0.01** |
| TLRNet | 0.97±0.59 | 1.09±0.78 | 0.21±0.44 | 0.24±0.45 |
| T-Learner | 1.68±1.88 | 2.22±3.15 | 0.12±0.10 | 0.28±0.57 |
| GANITE | 1.9±0.4 | 2.4±0.4 | 0.43±0.05 | 0.49±0.05 |
| DR-Learner | 2.38±2.9 | 2.63±3.64 | 0.14±0.14 | 0.32±0.59 |
| X-Learner | 2.89±3.93 | 3.07±4.42 | 0.25±0.35 | 0.40±0.69 |
| S-Learner | 3.18±4.90 | 3.30±5.31 | 0.43±0.91 | 0.55±1.21 |
| R-Learner | 4.01±4.2 | 3.76±4.34 | 0.43±0.66 | 0.52±1.06 |

$$\epsilon_{ATE} = \left| \frac{1}{n} \sum_{i=1}^{n} f(\varphi_t(x_i), 1) - f(\varphi_c(x_i), 0) - \frac{1}{n} \sum_{i=1}^{n} m_1(x_i) - m_0(x_i) \right| \quad (15)$$

While TARNet, Dragon-Net, and CFRNet produce impressive results, it is crucial to understand that these networks use specialized heads to estimate errors for each treatment group separately, making them more complex. In contrast, TLRNet achieves comparable results with a single, efficient estimator. TLRNet offers two main advantages over its competitors. First, it provides practical solutions using only a single estimator while closely matching the performance of the more complex and state-of-the-art methods. Second, it categorizes the treatment variable into two distinct informative spaces for a more straightforward interpretation.

## V. CONCLUSION

According to the results, the type of data used by the T-Learner method tends to be better estimated than that of the S-Learner method. However, the proposed TLRNet method utilizes a single estimator and has produced superior outcomes compared to the T-Learner method. Incorporating a target regulator is expected to enhance the results and introduce a new vision to the estimation and modeling processes in treatment effect research.

One of the major challenges of the TLRNet method is its estimator's variance compared to other deep learning approaches. This issue arises from using an estimator to evaluate the output at two treatment levels (representation functions). When the treatment effect is zero, we expect the representation of a sample, based on the representation space of both the treatment and control groups, to be similar. However, when the treatment effect is non-zero, we anticipate that different confounding features between the treatment and control groups will become more pronounced. As a result, when using a dataset from a randomized controlled trial with varying treatment effects, TLRNet may encounter difficulties in accurately predicting outcomes. We plan to target regularization in future work to better account for the treatment effect.